\documentclass[preprint,12pt]{elsarticle}
\usepackage{graphicx}
\usepackage{amssymb}
\usepackage{lineno}

\usepackage{hyperref}

\journal{Traffic4cast Competition by IARAI}

\begin{document}

\begin{frontmatter}


\title{Traffic4cast--Traffic Map Movie Forecasting\\Team MIE-Lab}



\author{Henry Martin\textsuperscript{1}, Ye Hong\textsuperscript{1}, Dominik Bucher\textsuperscript{1}, \\Christian Rupprecht\textsuperscript{2}, Ren\'{e} Buffat\textsuperscript{1}}

\address{\textsuperscript{1}Institute of Cartography and Geoinformation, ETH Zurich,\\Stefano-Franscini-Platz 5, 8093 Zürich, Switzerland\\\textsuperscript{2}Visual Geometry Group, University of Oxford,\\25 Banbury Rd, Oxford OX2 6NN, Great Britain}

\begin{abstract}
The goal of the IARAI competition \textit{traffic4cast} was to predict the city-wide traffic status within a 15-minute time window, based on information from the previous hour. 
The traffic status was given as multi-channel images (one pixel roughly corresponds to 100x100 meters), where one channel indicated the traffic volume, another one the average speed of vehicles, and a third one their rough heading. 
As part of our work on the competition, we evaluated many different network architectures, analyzed the statistical properties of the given data in detail, and thought about how to transform the problem to be able to take additional spatio-temporal context-information into account, such as the street network, the positions of traffic lights, or the weather.
This document summarizes our efforts that led to our best submission, and gives some insights about which other approaches we evaluated, and why they did not work as well as imagined. 
\end{abstract}

%\begin{keyword}
%Science \sep Publication \sep Complicated


%\end{keyword}

\end{frontmatter}

\section{Introduction}
The Institute of Advanced Research in Artificial Intelligence (IARAI) is a recently-founded research institute that organized the \textit{traffic4cast} competition together with the company HERE Technologies\footnote{See also \href{https://www.iarai.ac.at/traffic4cast/}{www.iarai.ac.at/traffic4cast}, \href{https://www.iarai.ac.at}{www.iarai.ac.at} and \href{https://www.here.com}{www.here.com}.}.
The \textit{traffic4cast} competition revolves around the prediction of traffic status in three major cities: Berlin, Istanbul and Moscow.
The recorded traffic was aggregated into 100x100 meters bins and made available as three-channel images.
Within these images, the first channel depicts the traffic volume in each cell, the second one the average speed of vehicles, and the third one the majority of vehicles' directions (as one of four cardinal directions).
The data spanned a whole year in 5-minute intervals, where certain days were left out from the training data, to be used for prediction and upload to the traffic4cast servers, which then would assess the quality of the prediction.
The prediction itself consisted of ``three images into the future'' (spanning a 15-minute interval), based on the previous hour (12 images).

Given the problem formalization, our efforts mostly focused on the application of well-known image processing algorithms, though we also explored various simple baselines, neural networks taking into account spatio-temporal context, as well as more complex network architectures that should be able to take advantage of the fact that the origin of the data stems from probes that move on a known graph.
Ultimately, we did not manage to outperform the ``simple'' application of a widely-used image processing algorithm, which might be a hint that either a lot more research on networks targeted specifically at this problem or a different formulation of the problem altogether is required.
Our best-performing network implementation can be found under \href{https://github.com/mie-lab/traffic4cast}{github.com/mie-lab/traffic4cast}, and is explained in the following sections.

\section{Data preparation}
The data preparation steps are implemented in the \emph{trafic4cast\_dataset} PyTorch dataloader class available in \textit{videoloader.py}. 
Before training our models on the data, we store the data uncompressed and re-transform it such that every 1-day traffic movie file has the shape $(t, c, h, w)$ with $t=288,\; c=3,\; h=495,\; w=436$. 
Furthermore we store the data block-wise using the h5py \emph{chunks} option with the frame-size as chunk-size. These preprocessing steps give a substantial speed-up when loading the files later. 
During the training and evaluation process we load clips of 15 sequential images and divide them into 12 training frames and 3 test frames. 
In order to be able to use non-LSTM based CNNs we reduce the dimensions of the data by collapsing the time dimension into the channel dimension. 
After this operation, the shape of the data is $(c*t, h, w)$, which results in a shape of $(36, 495, 436)$ for a single training sample.

For the evaluation of alternative methods the \emph{trafic4cast\_dataset} class has some additional optional features such as returning a vector with temporal features, returning the the city name per clip, sub-sample the data to train on a smaller portion of the image and it furthermore allows to train or evaluate only on the timestamps that are present in the test data.

\section{Method}
\begin{figure}[tbh]
\centering
\includegraphics[width=0.8\textwidth]{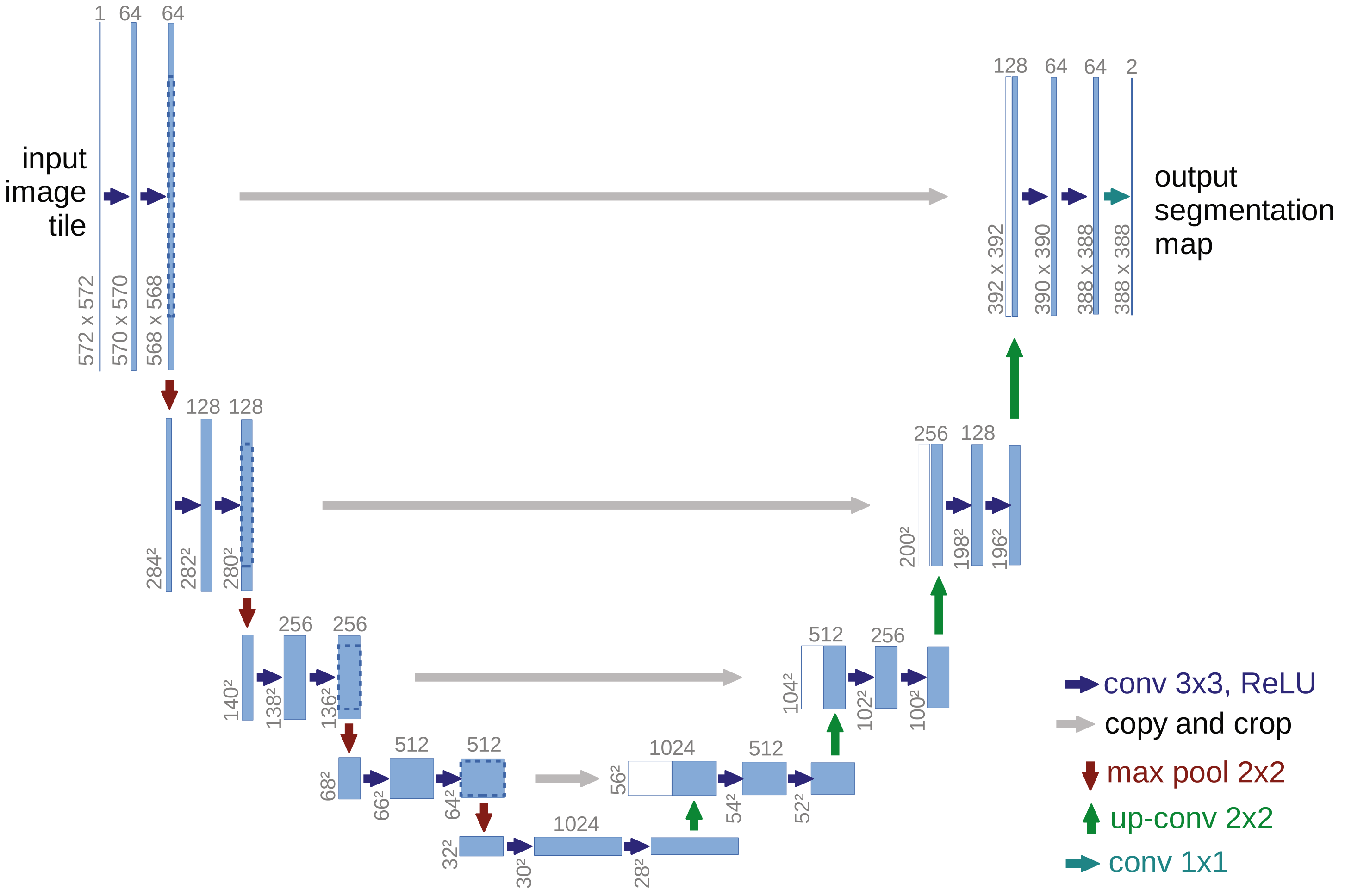}
\caption{The U-Net architecture from \protect \cite{ronneberger2015u}.}
\label{fig:Unet}
\end{figure}

In order to be able to use a regular CNN architecture, we use the above described data transformation to reshape all 12 training timestamps into the channel dimension. 
Using this data transformation in combination with regular CNNs allows us to greatly reduce the training and tuning complexity given by LSTM based approaches.
For our traffic4cast submission that reached the best score, we trained a single U-Net per city similar to the one shown in figure~\ref{fig:Unet} that was used in \cite{ronneberger2015u}. 

%%%% UNEt figure

The U-Nets all had depth 5, max-pooling for down sampling and a double-convolution in each layer. 
During training, we monitor the loss on the validation set as well as the loss on only the test times of the validation set. 
We use a batch-size of 5, stochastic gradient descent with a learning rate of 0.02, a momentum of 0.9, Nesterov momentum and reduce the learning rate to 0.001 after 5 epochs. In the end we clamp the results to values between 0 and 255. 

\section{Results}
\begin{figure}[htb]
\centering
\includegraphics[width=0.5\textwidth]{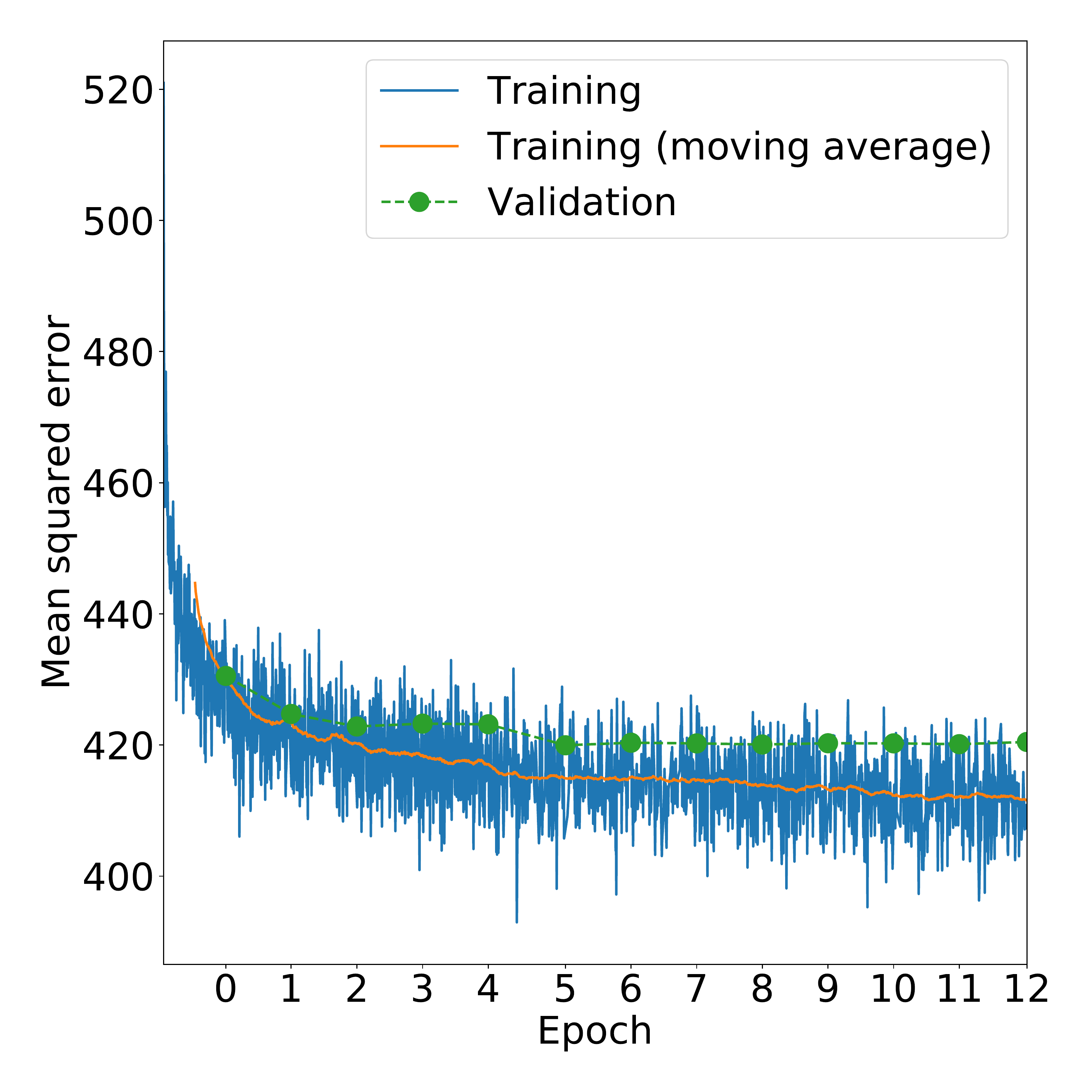}
\caption{The error of Berlin for a 12-epoch training run.}
\label{fig:loss}
\end{figure}

The training progress can be seen in figure~\ref{fig:loss}; here we use Berlin as an example. The training as well as the validation loss saturate relatively quickly after only few epochs. We see a small change in validation loss after the learning rate dropped but no further progress afterwards. 

%%%% Training figure

Ultimately, this led to a mean squared error (MSE) of $591.988$, though this varies enormously between cities (Berlin approx. 420, Istanbul 550, and Moscow 750) and between channels (volume approx. 50, speed 250 and heading 1000 for Berlin).
The reason for this is rather simple: as the heading is encoded into four different classes spanning the whole range from 0 to 255, wrong predictions tend to be penalized a lot more (than for example volume, which is usually close to zero for the majority of pixels). 
This can also easily be seen by taking a good baseline, which already reaches very low MSE errors for the volume and speed channels.
Similarly, the Moscow extent has many more streets than the other two extents, and the error scales with the number of pixels covered by a street (= number of non-zero pixels) , leading to a higher MSE for the Moscow predictions.
Finally, it should be noted that the three different predicted images (5-15 minutes into the future) are often very similar, this might be a hint, that regular CNN models are not not able to get a good ``sense of time'' but instead predict a likely ``average'' for all three time slots.

\section{Evaluated Alternative Methods}
We originally started with several different baselines, of which the best reached an MSE of $707.054$.
The best performing baseline filtered the training data to only contain the same time slots as in the test data, and used the average of those time slots over the year to make its prediction.
We then evaluated different network architectures, such as a (convolutional) LSTM-based image prediction network \cite{xingjian2015convolutional}, a deep lab \cite{chen2017deeplab}, a mask-based graph convolution neural network, a graph network that overlayed the images onto an OpenStreetMap street graph, a partial convolutional U-Net \cite{liu2018image}, a conditional U-Net, a ResNet \cite{he2016deep}, an ensemble model incorporating several of the others, and of course the vanilla U-Net \cite{ronneberger2015u} that led to our best prediction.
We also evaluated the effects of various masks (built from the pixels that never exceeded a certain threshold or always stayed zero), different optimizers, and a wide range of hyperparameters for each network.
The MSE of the predictions of all these models stayed in the range of $591.988$ to $707.054$.

\section{Conclusions}
The \textit{traffic4cast} competition gave us the chance to apply image processing algorithms and machine learning methods to movement and mobility data.
This was particularly interesting as it is still an actively researched field where it is not yet clear which approaches will lead to fruitful results.
The fact that one of the more simple approaches that we evaluated led to the best predictions might be a hint that either a lot more research towards very problem-specific network architectures or a different formulation of the problem altogether is required.
In our follow-up work, we intend to dig deeper into the problem and experiment with different formalisms that allow us to more easily incorporate natural constraints of the problem, such as the restriction to move on roads, or the interplay between different traffic streams at intersections, within and in between different areas of a city, etc.

\section*{Acknowledgments}
This research was supported by the Swiss Data Science Center (SDSC) and by the Swiss Innovation Agency Innosuisse within the Swiss Competence Center for Energy Research (SCCER) Mobility. Christian Rupprecht is supported by  ERC IDIU-638009.

% REFERENCES FORMAT
% References must be the same font size as other body text.
\bibliographystyle{SIGCHI-Reference-Format}
\bibliography{sample}

\end{document}